\documentclass[11pt,a4paper]{article}
\usepackage[hyperref]{eacl2021}
\usepackage{times}
\usepackage{latexsym}

\usepackage{url}
\usepackage{hyperref}
\usepackage{amsmath}
\usepackage{amsfonts}
\usepackage{multirow}
\usepackage{booktabs}
\usepackage{graphicx}
\usepackage{subcaption}
\usepackage{commath}
\usepackage{lipsum}
\usepackage{cleveref}
\usepackage{hyperref}

\usepackage{microtype}
\usepackage{mathtools, cuted}

\DeclarePairedDelimiter\ceil{\lceil}{\rceil}
\aclfinalcopy 

\title{SEMIE: SEMantically Infused Embeddings with Enhanced Interpretability for Domain-specific Small Corpus}

\author{Rishabh Gupta \and Rajesh N Rao\\
	Research and Technology Center, Robert Bosch, Bengaluru, India \\
	\texttt{\{Gupta.Rishabh, RajeshNagaraja.Rao\}@in.bosch.com} \\}

\date{}

\begin{document}
\maketitle
\begin{abstract}
Word embeddings are a basic building block of modern NLP pipelines. Efforts have been made to learn rich, efficient, and interpretable embeddings for large generic datasets available in the public domain. However, these embeddings have limited applicability for small corpora from specific domains such as automotive, manufacturing, maintenance and support, etc. In this work, we present a comprehensive notion of interpretability for word embeddings and propose a novel method to generate highly interpretable and efficient embeddings for a domain-specific small corpus. We report the evaluation results of our resulting word embeddings and demonstrate their novel features for enhanced interpretability.  

%
\end{abstract}

\section{Introduction}
Distributed representations of words, also termed as word embeddings, have been used extensively to excel at various applications such as parsing~\cite{lazaridou2013fish, bansal2014tailoring}, named entity recognition~\cite{guo2014revisiting}, image captioning~\cite{you2016image} and sentiment analysis~\cite{socher2013recursive}. They have also proven effective in modeling cognitive operations such as the judgement of word similarity~\cite{turney2010frequency, baroni2010distributional}, and the brain activity elicited by specific concepts~\cite{mitchell2008predicting}. 
However, these representations contain mappings of words to vectors of real numbers in dense and continuous space, and thus, inherently difficult to interpret. 

Recent studies~\cite{murphy2012learning, fyshe2014interpretable} suggest that sparsity and non-negativity of the word embeddings are two important characteristics that make them interpretable. The sparsity makes each word vector contain a small number of active (non-zero) dimensions~\cite{olshausen1997sparse}, which helps in increasing their separability and stability in the presence of noise~\cite{lewicki2000learning, donoho2005stable}.
In addition, the studies define the notion of interpretability in terms of the coherence of dimensions of the word embeddings~\cite{faruqui2015sparse, lipton2018mythos, subramanian2018spine}. In other words, the word embeddings are considered as interpretable if their dimensions denote specific semantic concepts. However, these studies are primarily focused on pre-trained word embeddings like GloVe~\cite{pennington2014glove} and word2vec~\cite{mikolov2013distributed}. These pre-trained embeddings are generated using millions of documents from generic public domain datasets such as Wikipedia and Google News, which contain billions of words. Also, to interpret the thousands of dimensions of the \textbf{s}parse \textbf{n}on-\textbf{n}egative (hereafter \textbf{`SNN'}) word embeddings\footnote{Usually the SNN word embeddings have at least 10x-times more dimensions as compared to the dense word embeddings.~\cite{faruqui2015sparse, subramanian2018spine}} and understand the sense they correspond to, we still need human judges who put in manual efforts and read these unlabeled dimensions.

In addition to the aforementioned semantic similarities, the recent studies suggest that the interpretability of word embeddings should also consider the semantic dissimilarities such as identification of discriminative word triples~\cite{krebs2016capturing, krebs2018semeval} . Given a triplet of words ($w_1$, $w_2$, $d$), the word embeddings should be able to determine whether $d$ is a discriminative feature between two concepts $w_1$ and $w_2$. For example, the word ``buckle'' is a discriminative feature in the triplet 
(``seat belt'', ``tires'', ``buckle'') that characterizes the first concept but not the second. 
Researchers have formulated this property as a binary classification task and proposed machine learning and similarity-based methods to evaluate the word embeddings ~\cite{zhang2018umd, dumitru2018alb, grishin2018igevorse}. However, to perform these evaluations for a domain-specific small corpus, we would need a manually curated set of discriminative (positive) and non-discriminative (negative) triples, which can be costly and time-consuming to curate. 

\emph{Comprehensive Notion of Interpretability}: In conclusion, we can say that the interpretability of word embeddings is expressed in terms of both semantic similarities and dissimilarities while representing them in SNN embedding space. Here, semantic similarities correspond to dimensional coherence, whereas semantic dissimilarities refer to properties such as identification of discriminative word triples. This raises the following question:\\  

\emph{How can we generate word embeddings for domain-specific small corpus, which are interpretable in terms of both semantic similarities and dissimilarities, when represented in SNN embedding space?}\\

To address the above-mentioned question, we present a novel method to generate such word embeddings, which we name as SEMantically Infused Embeddings (SEMIE). We showcase the enhanced interpretability (both in terms of semantic similarities and dissimilarities) capabilities of SEMIE while representing them in SNN embedding space. We also demonstrate the efficiency of SEMIE on the downstream classification task, both in dense and SNN embedding space.

\section{Preliminaries}
In this section, we discuss following preliminaries based on which we build our method to generate the SEMantically Infused Embeddings (SEMIE):

\begin{enumerate}
	\item A Semantic Infusion technique~\cite{gupta2020towards}, which helps in leveraging the meta-data of the text corpus to bring in semantic knowledge within the word embeddings. 
	
	
	\item A mathematical framework~\cite{yin2018dimensionality}, which helps in determining the optimal dimension of the word embeddings for a given text corpus. Such that these embeddings are expressive enough to capture all possible word relations while avoiding over-fitting.  
\end{enumerate}

\subsection{Semantic Infusion Technique}
\label{subs:sem_inf}
Semantic Infusion~\cite{gupta2020towards} is an efficient technique to associate meta-data within the text corpus. Using this, we can infuse special markers, referred to as \emph{Anchors}, within each document of the corpus. Given a document $d_i$ of length $l_{d_i}$ and belonging to a category class $c_j$ in the corpus, an anchor term $A_{c_j}$ is infused at $I_{freq}$ random and non-consecutive positions within the document. Here, $I_{freq}$ denotes \emph{Infusion Frequency} and its value is computed as: 

\begin{equation}
\label{eq:inf_freq}
I_{freq} = \ceil*{\frac{\log_2(l_{d_i})}{2}}
\end{equation}

For e.g., a sentence \emph{``new procurement scheme for farmers to focus on all crops.''} of a document class $c_j = $  \emph{India} will be processed as follows:

%

\begin{center}
	\emph{``new \textbf{A\_India} procurement scheme \textbf{A\_India} for farmers to focus on all \textbf{A\_India} crops.''} 
\end{center}

Intuitively, when an infused corpus like this is used to generate the word embeddings, the anchor terms help in identifying the semantically significant words based on the co-occurrence statistics.

\subsection{Optimal Dimensionality}
\label{subs:opt_dim}
The selection of dimensionality hyper-parameter is critical for the efficiency of any word embeddings. Researchers select dimensionality either in an ad hoc manner or using a grid search. Otherwise, they choose $300$, which is the most commonly used dimensionality while generating the word embeddings using millions of documents~\cite{mikolov2013distributed, pennington2014glove, bojanowski2017enriching}. 
But, for a domain-specific small corpus, we need a precise optimal dimensionality, otherwise it may lead to sub-optimal performance. 

To identify the optimal dimensionality, we leverage the mathematical framework as proposed in~\cite{yin2018dimensionality}. It determines the optimal dimensionality $d$, where $d \le k$, as the one which minimizes the Pairwise Inner Product (PIP) loss between an oracle embeddings $E \in \mathbb{R}^{n \times k}$ and the trained embeddings $\hat{E} \in \mathbb{R}^{n \times d}$ as defined by:

\begin{equation}
\begin{aligned}
\label{eq:pip}
\norm{\text{PIP}(E) - \text{PIP}(\hat{E})} = {} \norm{EE^T - \hat{E}\hat{E}^T}\\ = k - d + 2\norm{\hat{E}^TE^{\bot}}^2
\end{aligned}
\end{equation}

\begin{figure*}[]
	\includegraphics[width=\linewidth]{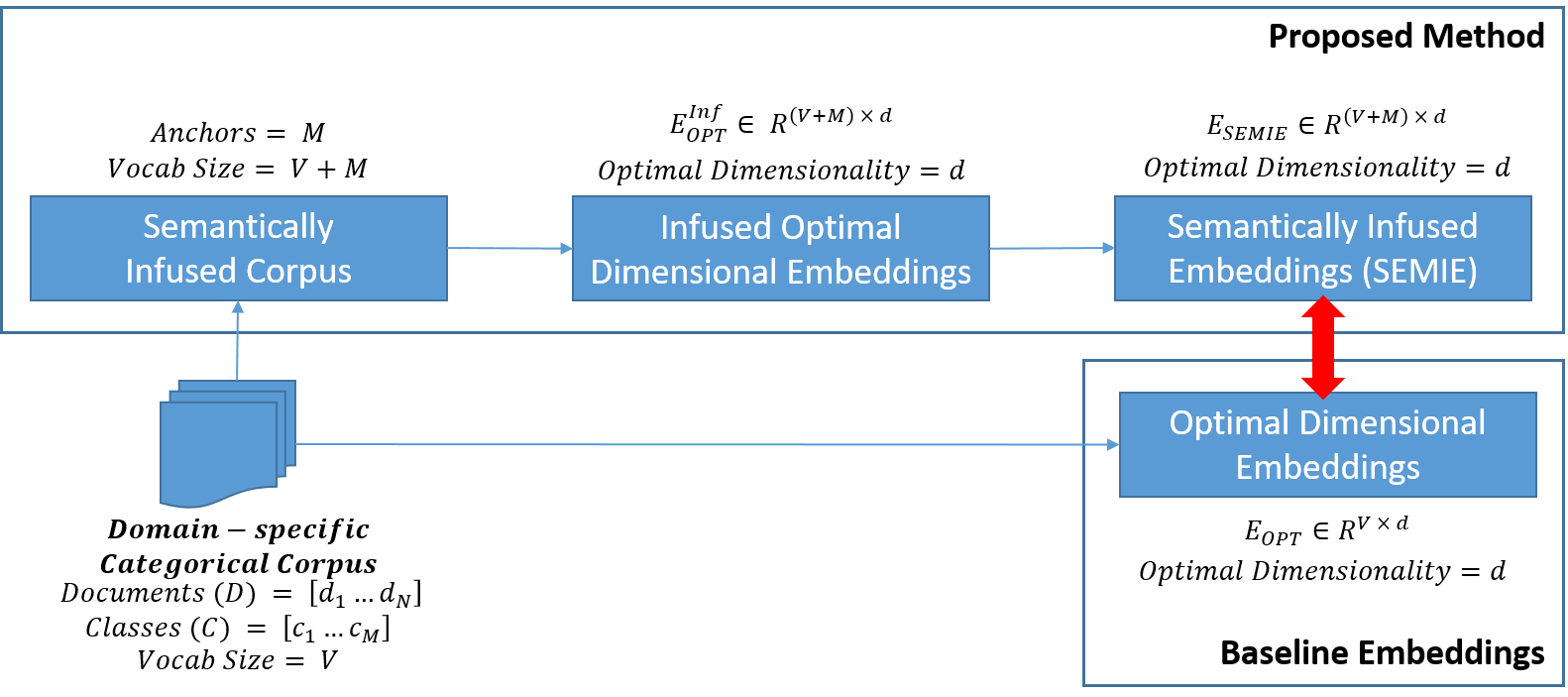}
	\caption{Flowchart for the proposed method. Given a domain-specific categorical corpus $\mathcal{D}$, we first generate the semantically infused corpus using the semantic infusion technique as explained in the Section~\ref{subs:sem_inf}. Next, using a mathematical framework as explained in the Section~\ref{subs:opt_dim}, we compute the optimal dimensionality $d$ for the infused corpus and generate the infused optimal dimensional embeddings $E^{Inf}_{OPT}$ using word2vec. Last, we generate the semantically infused embeddings $E_{SEMIE}$ as described in Section~\ref{sec:method}. In addition, we also generate the baseline optimal dimensional embeddings $E_{OPT}$ which we use to evaluate $E_{SEMIE}$ in terms of Interpretability and downstream classification task performance.}
	\label{fig:flowchart}
\end{figure*}

We use this framework as an intermediatory step to compute the optimal dimensionality of the corpus while generating SEMIE, as shown in Figure~\ref{fig:flowchart}.

%

	
%

%


%

%

\pagebreak
\section{Methodology}
\label{sec:method}
In this section, we describe the proposed method used to generate SEMIE, as illustrated in Figure~\ref{fig:flowchart}. Given a domain-specific categorical corpus of $N$ documents, $\mathcal{D} = [d_1 \dots d_N]$, where each document $d_i$ belongs to a category class $c_j$, in the set of $M$ classes, $\mathcal{C} = [c_1 \dots c_M]$.  We generate the semantically infused embeddings $E_{SEMIE} \in \mathbb{R}^{(V+M) \times d}$,  where $V$ is the vocabulary size of the corpus, $M$ are the anchor terms infused using the semantic infusion technique, and $d$ is the computed optimal dimensionality of word embeddings of the corpus. The key steps of the method are as follows:

\begin{enumerate}
	\item \textbf{Semantically Infused Corpus}
	
	 We take the domain-specific categorical corpus $\mathcal{D}$ and generate the semantically infused corpus using the semantic infusion technique, as explained in Section ~\ref{subs:sem_inf}. After this step, the vocabulary size of the entire corpus increases from $V$ to $(V + M)$, where $V$ is the vocabulary size of the initial corpus, and $M$ are the infused additional anchor terms. 
	
	\item \textbf{Optimal Dimensional Embeddings} 
	
	We take the semantically infused corpus, compute the optimal dimensionality $d$, as explained in Section~\ref{subs:opt_dim} and generate the infused optimal dimensional embeddings $E^{Inf}_{OPT} \in \mathbb{R}^{(V+M) \times d}$ using the word2vec method as demonstrated in the Figure~\ref{fig:flowchart}. 
	
	
	\item \textbf{Semantically Infused Embeddings} 
	
	We take the infused optimal dimensional embeddings $E^{Inf}_{OPT} \in \mathbb{R}^{(V+M) \times d}$ and generate the semantically infused embeddings $E_{SEMIE} \in \mathbb{R}^{ (V+M) \times d}$ as follows:
	\linebreak
	
	For each column $C_i$ of the embeddings matrix $E^{Inf}_{OPT}$, we sort the column in the increasing order of values and then select the values of $M$ anchor terms. For each anchor term $A_{C_i}$ and non-anchor word $w_{C_i}$ pair in the column $C_i$, we compute a semantic weight $w_s$, as given in Equation~\ref{eq:sem_w} and add it to the value of non-anchor word $w_{C_i}$. This gives us the semantically infused embeddings $E_{SEMIE}$.
	
	\begin{equation}
	\label{eq:sem_w}
	w_s = \frac{A_{C_i}}{||index(A_{C_i}) - index(w_{C_i})||} 
	\end{equation}
	
	 Intuitively, this means that in each column (dimension) of the semantically infused embeddings $E_{SEMIE}$, the words in the neighborhood of the anchors will form semantically coherent groups such as top words in Table~\ref{tab:kwd}.
	
\end{enumerate}
%
%




\begin{table*}[ht]
	\caption{Detailed statistics of the three datasets used in this work. For each dataset, the optimal embedding dimension $d$ is computed using the mathematical framework as described in Section~\ref{subs:opt_dim}. The SNN embedding dimension is taken as  $10\times d$. The $\#$ Anchors are equal to the $\#$ classes in the dataset.}
	
	\label{tab:dataset}
	\begin{tabular}{l|c|c|c|c|c}
		\toprule
		
		\begin{tabular}[c]{@{}c@{}}Dataset\end{tabular} & \begin{tabular}[c]{@{}c@{}}\# \\ Docs\end{tabular} & \begin{tabular}[c]{@{}c@{}}\# \\ Anchors / Vocab\\ ($M$/ $V$)\end{tabular} & 
		\begin{tabular}[c]{@{}c@{}}Class Names\end{tabular}
		& \begin{tabular}[c]{@{}c@{}}Optimal \\ Dimension\\ ($d$)\end{tabular}
		& \begin{tabular}[c]{@{}c@{}}SNN \\ Dimension\\ ($10\times d$)\end{tabular} \\

		\midrule
		AG News                                                 & 938,076                                                 & 10 / 92,721                                                    & \begin{tabular}[c]{@{}c@{}}Business, Sci/Tech,\\ Entertainment,  Sports,\\ World, Top News, Europe,\\ Italia, U.S., Health\end{tabular}                                        & 103      & 1030                                                         \\
		\midrule
		NHTSA                                                    & 800,815                                                 & 10 / 45,762                                                   & \begin{tabular}[c]{@{}c@{}}Steering, Visibility,\\ Suspension, Structure,\\  Electrical System, Fuel System,\\  Air Bags, Seat Belts,  Tires\\ Vehicle Speed Control\end{tabular} & 113              & 1130                                                 \\
		\midrule
		\begin{tabular}[c]{@{}c@{}}Yahoo \\ Answers\end{tabular} & 730,000                                                 & 5 / 131,607                                                     & \begin{tabular}[c]{@{}c@{}}Science \& Mathematics, Health,\\ Sports, Business \& Finance,\\ Politics \& Government\end{tabular}                                                                     & 139   & 1390                                           \\
		\bottomrule
	\end{tabular}
\end{table*}

\section{Experimental Setup}
We study the efficacy of SEMIE in terms of performance in downstream classification task both in dense and SNN embedding space. Furthermore, we demonstrate the enhanced interpretability of SEMIE both in terms of semantic similarities and dissimilarities when transformed in SNN embedding space.

In the following subsections, we describe the datasets used and the generation of baseline optimal dimensional embeddings $E_{OPT}$ and semantically infused embeddings $E_{SEMIE}$ for each dataset. 
In addition, we describe two state-of-the-art methods namely SPINE~\cite{subramanian2018spine} and SPOWV~\cite{faruqui2015sparse} used to transform the $E_{OPT}$ and $E_{SEMIE}$ to SNN embedding space. Later, for each dataset, we present the effectiveness of $E_{SEMIE}$ on downstream classification task (Section~\ref{sec:classi}) and interpretability in terms of semantic similarities (Section~\ref{sec:interpret}) and dissimilarities (Section~\ref{sec:kwc}) as compared to $E_{OPT}$.



\subsection{Datasets}
In our experiments, we use $3$ datasets: $(1)$ \href{https://www.nhtsa.gov/}{NHTSA} (National Highway Traffic Safety Administration), a web forum platform, which contains complaints of the automobile domain registered by the consumers. $(2)$ \href{https://in.answers.yahoo.com/answer}{Yahoo} answers, a question-answering platform, which contains answers to various questions posted by online users. $(3)$ \href{http://groups.di.unipi.it/~gulli/AG\_corpus\_of\_news\_articles.html}{AG NEWS}, a collection of news articles gathered from more than $2000$ news sources. The statistics of datasets are summarized in the Table~\ref{tab:dataset}. 

\subsection{$E_{OPT}$ and $E_{SEMIE}$ Embeddings}
For each dataset of vocabulary size $V$, we compute the optimal dimensionality hyper-parameter $d$ as described in the Section~\ref{subs:opt_dim} and generate the baseline optimal dimensional embeddings $E_{OPT} \in \mathbb{R}^{V \times d}$ using the word2vec method. 
In addition, we generate the semantically infused corpus while infusing $M$ anchor terms and generate the semantically infused embeddings $E_{SEMIE} \in \mathbb{R}^{(V+M) \times d}$ for each dataset as explained in the Section~\ref{sec:method}.

In our experiments, we compare the $E_{SEMIE}$ embeddings w.r.t. the baseline $E_{OPT}$ embeddings for each dataset and demonstrate the effectiveness of $E_{SEMIE}$ in terms of downstream classification task performance and interpretability.

%

\begin{table*}[ht]
	\centering
	\caption{The effectiveness of SEMIE in terms of the downstream classification accuracy both in dense and SNN embedding space. $E_{SEMIE}$ either improves the classification accuracy or it is comparable to baseline $E_{OPT}$ across all datasets for both dense and SNN embedding space.}
	\label{tab:classification}
	\begin{tabular}{c|cc||cc|cc}
		\toprule
		\multirow{2}{*}{Dataset} & \multicolumn{2}{c||}{ Dense Embeddings } & \multicolumn{4}{c}{Sparse Embeddings}    
		\\
		\cmidrule{2-7}
		& $E_{OPT}$         & $E_{SEMIE}$      & \begin{tabular}[c]{@{}c@{}}$E_{OPT}$- \\ SPINE\end{tabular} & \begin{tabular}[c]{@{}c@{}}$E_{SEMIE}$- \\  SPINE\end{tabular} & \begin{tabular}[c]{@{}c@{}}$E_{OPT}$- \\  SPOWV\end{tabular} & \begin{tabular}[c]{@{}c@{}}$E_{SEMIE}$- \\  SPOWV\end{tabular}\\
		\midrule
		AG News                  & 76.95\%            & \textbf{78.10\%}   & 77.55\%                                                    & \textbf{78.28\%}                                                & 77.48\%                                                       & \textbf{77.93\%}    \\
		
		\midrule
		NHTSA                    & \textbf{87.45\%}    & \textbf{87.45\%}   & \textbf{87.48\%}                                             & \textbf{87.48\%}                                                & \textbf{88.63\%}                                              & 87.80\%  \\
		\midrule
		\begin{tabular}[c]{@{}c@{}}Yahoo \\Answers \end{tabular}                    & 75.47\%             & \textbf{76.09\%}   & 72.40\%                                                      & \textbf{73.18\%}                                                & 74.12\%
		
		& \textbf{74.84\%}\\
		\bottomrule                                          
	\end{tabular}
\end{table*}

\begin{table*}[]
	\centering
	\caption{The precision score comparisons on the word intrusion detection tests where higher precision numbers indicate more interpretable dimensions. $E_{SEMIE}$ improves the interpretability as compared to $E_{OPT}$ for NHTSA and Yahoo Answers datasets in SNN embedding space.}
	\label{tab:widt}
	\begin{tabular}{c|cc|cc}
		\toprule
		\multirow{2}{*}{Dataset} & 
		\multicolumn{4}{c}{Sparse Embeddings}    
		\\
		\cmidrule{2-5}
		&\begin{tabular}[c]{@{}c@{}}$E_{OPT}$-\\SPINE\end{tabular} & \begin{tabular}[c]{@{}c@{}}$E_{SEMIE}$-\\SPINE\end{tabular} & \begin{tabular}[c]{@{}c@{}}$E_{OPT}$-\\SPOWV\end{tabular} & \begin{tabular}[c]{@{}c@{}}$E_{SEMIE}$-\\SPOWV\end{tabular}\\
		\midrule
		AG News                  & 
		\textbf{32.67}                                                      & 30.00                                                & \textbf{21.67}                                                       & 18.00    \\
		
		\midrule
		NHTSA                    & 
		25.67                                             & \textbf{30.00}                                                & 29.67                                              & \textbf{30.00}  \\
		\midrule
		\begin{tabular}[c]{@{}c@{}}Yahoo \\Answers \end{tabular}                    & 
		24.33                                                    & \textbf{31.00}                                                & \textbf{34.00}                                                       
		& \textbf{34.00}\\
		\bottomrule                                          
	\end{tabular}
\end{table*}

\subsection{Sparse Non-negative Transformation}
To demonstrate the enhanced interpretability of SEMIE, we transform both $E_{OPT}$ and $E_{SEMIE}$ to SNN embeddings space for each dataset. We perform the transformations using $2$ state-of-the-art methods: $(1)$ SPINE: SParse Interpretable Neural Embeddings~\cite{subramanian2018spine} and $(2)$ Sparse Overcomplete Word Vector Representations (SPOWV) ~\cite{faruqui2015sparse}. Using SPINE and SPOWV, we transform both $E_{OPT}$ and $E_{SEMIE}$ embeddings of each dataset to a  $\mathbb{R}^{10\times d}$ space such that ($10\times d$)-dimensional embeddings are both sparse and non-negative. Here, $d$ is the computed optimal dimension for the word embeddings of the dataset as mentioned in the Table~\ref{tab:dataset}. 

\textbf{Hyper-parameters:} For SPINE and SPOWV, we use hyper-parameters which are best performing as per the authors’ recommendations. SPINE- No. of Epochs: $4000$, Sparsity: $0.85$, Noise: $0.2$; SPOWV- l1\_reg: $0.5$, l2\_reg: $1e-5$, num\_cores: $1$. 

After this step, we have baseline optimal dimensional embeddings: $E_{OPT}$, $E_{OPT}$-SPINE and $E_{OPT}$-SPOWV and semantically infused embeddings: $E_{SEMIE}$, $E_{SEMIE}$-SPINE and $E_{SEMIE}$-SPOWV for each dataset.

\section{Downstream Classification Task}
\label{sec:classi}
In this section, we evaluate the performance of SEMIE in the downstream classification task both in dense and SNN embedding space. For each dataset, we compare the classification accuracy while using $E_{SEMIE}$, $E_{SEMIE}$-SPINE and $E_{SEMIE}$-SPOWV w.r.t. $E_{OPT}$, $E_{OPT}$-SPINE and $E_{OPT}$-SPOWV respectively. 
%
We use the average of word vectors of the words in a document as features and experiment with SVMs for the text classification task. For each dataset, we take a balanced set of $1600$ documents as the training set and $400$ documents as the test set from each class of the dataset. We then combine the training sets and test sets and report the accuracy on the combined test set as listed in the Table~\ref{tab:classification}. 
Please note that for this evaluation we removed the $M$ anchor terms from all $E_{SEMIE}$ embedding matrices.

\begin{table*}[ht]
	
	\centering
	\caption{The Interpretation and labels of some sampled dimensions. For each dataset, we examine the top $5$ words and top-ranked anchor term for a few randomly chosen dimensions of  $E_{SEMIE}$-SPOWV SNN embeddings. Clearly, the word groups are interpretable and semantically coherent with the anchor terms.  Also, the anchor terms can be seen as the labels for the dimension as they are able to cater to the interpretations of the dimensions.}
	\label{tab:kwd}
	\begin{tabular}{c|c|c}
		\toprule
		Dataset                                                                  & \begin{tabular}[c]{@{}c@{}}Top-ranked Anchor in \\ Sampled Dimension\end{tabular} & \begin{tabular}[c]{@{}c@{}}Top Words in \\ Sampled Dimension\end{tabular} \\ \midrule
		\multirow{5}{*}{AG News}                                                 & A\_Health                                                                  & colon, murdoch, breast, smokers, older                                    \\ \cmidrule{2-3} 
		& A\_Europe                                                                  & holy, brighton, pope, xvi, benedict                                       \\ \cmidrule{2-3} 
		& A\_Sports                                                                  & touchdown, seconds, pointer, interceptions, tiebreaking                   \\ \cmidrule{2-3} 
		& A\_Top News                                                                & resumption, islamist, disarmament, zimbabwe, ceasefire                    \\ \cmidrule{2-3} 
		& A\_U.S.                                                                    & eliot, blitz, attorney, spitzer, buskirk                                  \\ \midrule
		\multirow{5}{*}{NHTSA}                                                   & A\_Electrical System                                                       & speakers, audio, radio, buttons, stereo                                   \\ \cmidrule{2-3} 
		& A\_Tires                                                                   & durability, cupped, engulfed, disintegrated, worn                         \\ \cmidrule{2-3} 
		& A\_Structure                                                               & attachment, creeping, framing, section, weakened                          \\ \cmidrule{2-3} 
		& A\_Seat Belts                                                              & webbing, anchorage, limiter, waist, unbuckle                              \\ \cmidrule{2-3} 
		& A\_Visibility                                                              & obstructing, hybrid, draining, contact, interfering                       \\ \midrule
		\multirow{5}{*}{\begin{tabular}[c]{@{}c@{}}Yahoo\\ Answers\end{tabular}} & A\_Health                                                                  & stones, mothers, hypothyroidism, thyroid, nmyth                           \\ \cmidrule{2-3} 
		& A\_Sports                                                                  & abuse, hyperactivity, psychiatric, drank, suffers                         \\ \cmidrule{2-3} 
		& A\_Business \& Finance                                                     & declined, angle, defects, worker, uninsured                               \\ \cmidrule{2-3} 
		& A\_Politics \& Government                                                  & peaceful, openly, dominion, altar, forefathers                            \\ \cmidrule{2-3} 
		& A\_Science \& Mathematics                                                  & longitude, language, cartesian, sapiens, coordinates                      \\ \bottomrule
	\end{tabular}
\end{table*}

\section{Interpretability}
\label{sec:interpret}
We evaluate the interpretability (in terms of semantic similarities) of SEMIE in SNN embedding space using the intrusion detection tests (Section~\ref{subs:widt}). These tests have widely been used in literature ~\cite{chang2009reading, murphy2012learning, fyshe2014interpretable, subramanian2018spine,faruqui2015sparse} to evaluate the interpretability of the word embeddings. 

Additionally, in SEMIE, due to the Anchor term infusion, we can get labels for the dimensions of SNN embeddings. This helps in better interpretation of the SNN embedding dimensions as the labels are coherent with the semantic concepts represented by the dimensions (Section~\ref{subs:kwd}).

%
%



\subsection{Word Intrusion Detection Test}
\label{subs:widt}

In a particular word intrusion detection test, a human judge is presented with $5$ words and asked to select the ``intruder'' or semantically the ``odd one out''. The strategy to generate these tests: for a given dimension (column) of the embeddings matrix, we select $4$ top-ranked words and a ``intruder'' word from the bottom half of the ranked list, which is also present in the top $10$ percentile in at least one other dimension. An example of the test is: ``hyundai'', ``crumbled'', ``spectra'', ``suzuki'', ``kia''. Here, ``crumbled'' is the intruder. 

For each dataset, we compare the precision scores on the word intrusion detection tests while using $E_{SEMIE}$, $E_{SEMIE}$-SPINE and $E_{SEMIE}$-SPOWV w.r.t. $E_{OPT}$, $E_{OPT}$-SPINE and $E_{OPT}$-SPOWV embeddings respectively. We randomly sample $100$ dimensions of all embedding matrices and generate the intrusion detection tests. In total, we generate $1800$ tests and get each test annotated independently by the $3$ judges. The results of these intrusion detection tests are listed in Table~\ref{tab:widt}. 
Please note that for this evaluation we removed the $M$ anchor terms from all $E_{SEMIE}$ embedding matrices.

%
%

\begin{figure*}[]
	
	\includegraphics[width=\linewidth]{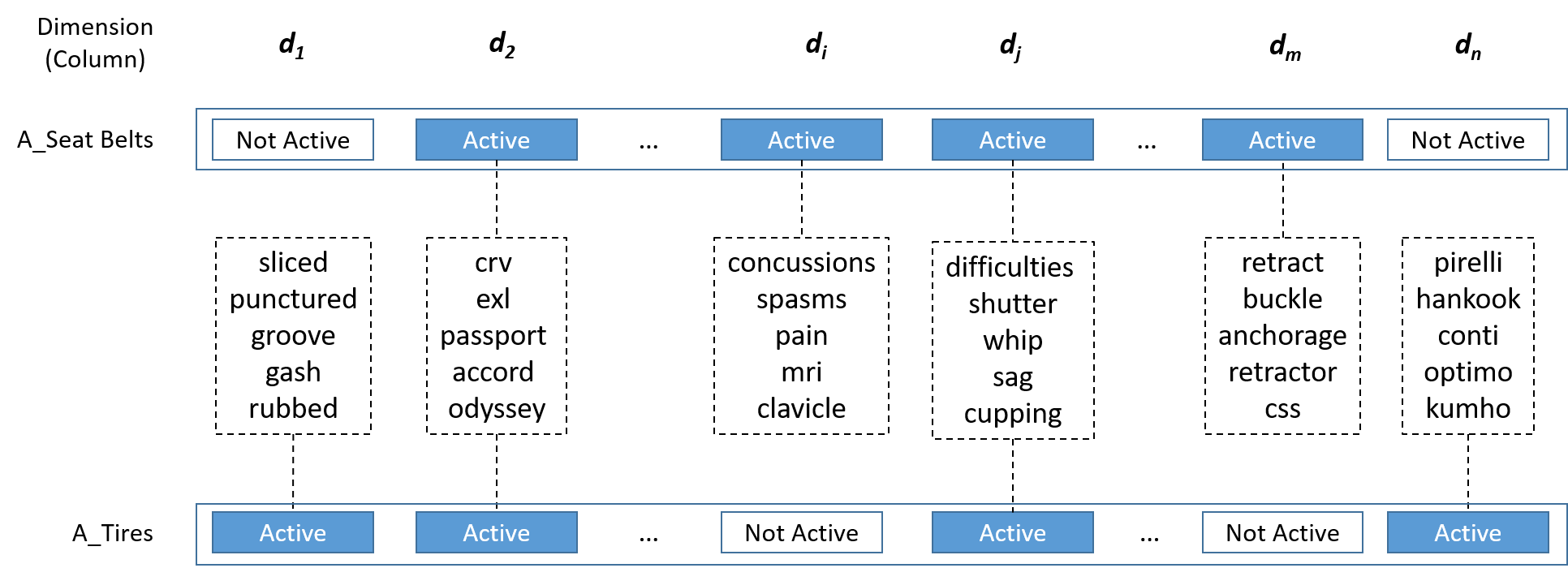}
	\caption{An illustration of the discriminative and non-discriminative features as identified using SEMIE. We examine the embedding vectors for the anchors ``A\_Seat Belts'' and ``A\_Tires'' in the $E_{SEMIE}$-SPINE embeddings of the NHTSA dataset. Clearly, the top words in the dimensions when only one anchor is active are discriminative features corresponding to the concepts Seat Belts and Tires, for example, the word``punctured'' (an issue specific to tires). This helps us in creating the discriminative triple such as \emph{(Tire, Seat Belts, punctured)}. Whereas, the top words in dimensions when both anchors are active (non-zero) are non-discriminative features such as the word ``crv'' (a car model name common across both seat belts and tires). This helps us in creating the non-discriminative triple such as \emph{(Tire, Seat Belts, crv)}.}
	\label{fig:kwc}
\end{figure*}

\subsection{Dimensions' Labels}
\label{subs:kwd}
In this section, we demonstrate that SEMIE provides a novel mechanism to automatically label SNN embedding dimensions. These labels are able to cater to the interpretations of the top words within the dimensions and thus help in a better understanding and interpretability of the dimensions.

We take the SNN semantically infused embeddings $E_{SEMIE}$-SPOWV for each dataset and randomly sample a few the dimensions (columns) of the embeddings matrix. For each sampled dimension, we sort the column in the decreasing order of values and then select top $5$ words and the top-ranked anchor term in that dimension as listed in the Table~\ref{tab:kwd}. 
Here, we observe that the semantic groups formed by the top words in these sampled dimensions are highly coherent with the top-ranked anchor term.


\subsection{Semantic Dissimilarities} 
\label{sec:kwc}

The interpretability of word embeddings is also expressed in terms of semantic dissimilarities such as identification of discriminative and non-discriminative word triples. A word triplet ($w_1$, $w_2$, $d$) is termed as discriminative if the feature $d$ characterizes the first concept $w_1$ but not the second concept $w_2$ and thus act as a discriminative feature between the two concepts as listed in Table~\ref{tab:dis}.


SEMIE provides a novel mechanism to identify the discriminative and non-discriminative triples in an unsupervised fashion. To showcase this, we take the $E_{SEMIE}$-SPINE SNN embeddings of the NHTSA dataset and select the vectors of ``A\_Seat Belts'' and ``A\_Tires'' anchor terms. For each dimension in both vectors, we then identify which one is active (non-zero) and which one is not active (zero), as illustrated in the Figure~\ref{fig:kwc}. 

We observe that the top five words for the dimensions where both anchors are active are non-discriminative features corresponding to the concepts ``Tire'' and ``Seat Belts'' in the dataset. For example, a mix of seat belt and tire issues or the common car model names. Thus, we can formulate the non-discriminative triples such as \emph{(Tire, Seat Belts, crv)} and \emph{(Tire, Seat Belts, passport)} related to the concepts ``Tire'' and ``Seat Belts''.

Whereas, the top five words for the dimensions where only one anchor is active are discriminative features such as specific tire manufacturing company names, issues related specifically to either tires or seat belts, and injuries related to seat belts. Thus, we can formulate the discriminative triples such as \emph{(Tire, Seat Belts, punctured)} and \emph{(Seat Belts, Tire, buckle)} related to the concepts ``Tire'' and ``Seat Belts''. The identified discriminative and non-discriminative triplets for the concepts ``Tire'' and ``Seat Belts'' are listed in the Table~\ref{tab:dis}.

\begin{table*}[]
	\centering
	\caption{The discriminative and non-discriminative triples. A triplet (Concept 1, Concept 2, Feature) is termed as discriminative if the ``Feature'' characterizes the concept ``Concept 1'' but not the concept ``Concept 2''. We examine the two concepts ``Seat Belts'' and ``Tires'' using $E_{SEMIE}$-SPINE embeddings of the NHTSA dataset as shown in Fig~\ref{fig:kwc}. Clearly, the ``Feature'' characterizes the ``Concept 1'' in the discriminative triples whereas in the non-discriminative triples the ``Feature'' are generic features corresponding to both ``Concept 1'' and ``Concept 2".}
	\label{tab:dis}
	\begin{tabular}{c||c}
		\toprule
		\multicolumn{1}{c||}{Discriminative Triples} & \multicolumn{1}{c}{Non-discriminative Triples}    \\
		\midrule
		(Tire, Seat Belts, sliced)      & (Tire, Seat Belts, crv)          \\
		\midrule
		(Tire, Seat Belts, punctured)   & (Tire, Seat Belts, exl)          \\
		\midrule
		(Tire, Seat Belts, groove)      & (Tire, Seat Belts, passport)     \\
		\midrule
		(Tire, Seat Belts, gashed)      & (Tire, Seat Belts, accord)       \\
		\midrule
		(Tire, Seat Belts, rubbed)      & (Tire, Seat Belts, odyssey)      \\
		\midrule
		(Seat Belts, Tire, retract)     & (Tire, Seat Belts, difficulties) \\
		\midrule
		(Seat Belts, Tire,  buckle )     & (Tire, Seat Belts, shutter)      \\
		\midrule
		(Seat Belts, Tire, anchorage )  & (Tire, Seat Belts, whip)         \\
		\midrule
		(Seat Belts, Tire,  retractor )  & (Tire, Seat Belts, sag)          \\
		\midrule
		(Seat Belts, Tire,  css )        & (Tire, Seat Belts, cupping)  \\
		\bottomrule   
	\end{tabular}
\end{table*}

\section{Results and Discussions}
In this section, we summarize the results of our experiments and discuss their benefits and impacts.\\

\noindent \textbf{Downstream Classification Task Performance} 

Based on the experiment performed in Section~\ref{sec:classi} and the results shown in Table~\ref{tab:classification}, it is evident that SEMIE's performance is competitive as compared to the baseline $E_{OPT}$ across all domain-specific small datasets for both in dense and SNN embedding space.\\

\noindent \textbf{Interpretability: Semantic Similarities} 

Based on precision accuracies in the word intrusion detection tests, as shown in Table~\ref{tab:widt}, it is clear that SEMIE's performance is comparable to the baseline $E_{OPT}$ across majority of the datasets. This implies that, for domain-specific small datasets, SEMIE is an interpretable word embedding representation when transformed in SNN embedding space. \\

\noindent From the Table~\ref{tab:kwd}, it is evident that SEMIE provides a novel mechanism to provide labels for SNN embedding dimensions. These labels are able to cater to the interpretations of the embedding dimensions and thus help in quick and easy understanding.\\

\noindent \textbf{Interpretability: Semantic Dissimilarities} 

Based on the demonstration shown in Section~\ref{sec:kwc} and Figure~\ref{fig:kwc}, we observe that, given two concepts, SEMIE provides a novel mechanism to generate the discriminative and non-discriminative words triples from SNN embeddings. These triples can help in the better assessment of the embeddings' interpretability in terms of semantic dissimilarities.

\section{Conclusion}
We define a comprehensive notion of interpretability of word embeddings which includes both semantic similarities and dissimilarities while representing them in SNN embedding space. We present a novel method to generate efficient and highly interpretable word embeddings (SEMIE) for a domain-specific small corpus. Also, we demonstrate that SEMIE provides novel mechanisms to provide labels for SNN embedding dimensions and to generate discriminative/non-discriminative word triples in an unsupervised fashion. 


%

\bibliography{anthology,eacl2021}
\bibliographystyle{acl_natbib}

\end{document}